\documentclass[sigplan,nonacm]{acmart}

\graphicspath{ {./fig/} }
\usepackage{units} 
\usepackage{subcaption} 
\usepackage{varioref} 
\usepackage{listings} 
\usepackage{cancel} 

\newcommand{\cross}{\times} 
\newcommand{\real}{\mathbb{R}} 
\newcommand{\degree}{\text{\textdegree}} 
\allowdisplaybreaks[1]

\newcommand{\noun}[1]{\textsc{#1}} 
\newcommand{\textbit}[1]{\textbf{\textit{#1}}} 

\theoremstyle{remark}
	\newtheorem*{rem}{Remark}

	\newtheorem*{problem*}{Problem}
	\newtheorem*{prop*}{Proposition}
	
	\newtheorem*{example*}{Example}
\theoremstyle{definition}

\theoremstyle{plain}
	\newtheorem{thm}{Theorem}

\lstdefinelanguage{GLSL}
{
	morekeywords={
		attribute,const,uniform,varying,
		break,continue,do,
		for,while,
		switch,case,default,
		if,else,
		in,out,inout,
		float,int,void,bool,true,false,
		discard,return,
		mat2,mat3,mat4,
		mat2x2,mat2x3,mat2x4,
		mat3x2,mat3x3,mat3x4,
		mat4x2,mat4x3,mat4x4,
		vec2,vec3,vec4,
		ivec2,ivec3,ivec4,
		bvec2,bvec3,bvec4,
		uint,uvec2,uvec3,uvec4,
		lowp,mediump,highp,precision,invariant,
		sampler1D,sampler2D,sampler3D,samplerCube,
		struct,
		define,undef,
		ifdef,ifndef,elif,endif,
	},
	sensitive=false, 
	morecomment=[l]{//},
	morecomment=[s]{/*}{*/},
	morestring=[b]"
}
\received{April 2020} 
\received{May 2020} 
\received[this version]{May 2023}

\begin{document}

\title{Calculating Pose with Vanishing Points of Visual-Sphere Perspective Model}

\author{Jakub Maksymilian Fober}
\orcid{0000-0003-0414-4223}
\email{talk@maxfober.space}

\renewcommand\shortauthors{Fober, J.M.}

\begin{abstract}
	The goal of the proposed method is to directly obtain a pose matrix of a known rectangular target, without estimation, using geometric techniques. This method is specifically tailored for real-time, extreme imaging setups exceeding 180\textdegree\ field of view, such as a fish-eye camera view.
	The introduced algorithm employs geometric algebra to determine the pose for a pair of coplanar parallel lines (ideally a tangent pair as in a rectangle). This is achieved by computing vanishing points on a visual unit sphere, which correspond to pose matrix vectors. The algorithm can determine pose for an extremely distorted view source without prior rectification, owing to a visual-sphere perspective model mapping of view coordinates. Mapping can be performed using either a perspective map lookup or a parametric universal perspective distortion model, which is also presented in this paper.
	The outcome is a robust pose matrix computation that can be executed on an embedded system using a microcontroller, offering high accuracy and low latency.
	This method can be further extended to a cubic target setup for comprehensive camera calibration. It may also prove valuable in other applications requiring low latency and extreme viewing angles.
\begin{figure}[h]
	\includegraphics{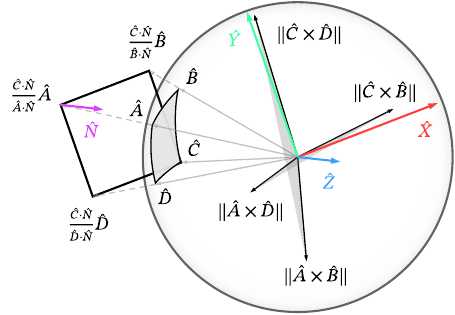}
	\caption[3D rectification and pose determination]
	{3D rectification and pose determination of $\overline{ABCD}$ quad using vanishing points on a visual-sphere model.}
\end{figure}
\end{abstract}




\maketitle


\begin{figure}[bh]
	\footnotesize
	\copyright\ 2020 Jakub Maksymilian Fober\smallskip\\
	\href{https://creativecommons.org/licenses/by-nc-nd/3.0/}
		{\includegraphics{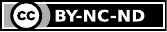}}\\
	This work is licensed under Creative Commons BY-NC-ND 3.0 license.
	\url{https://creativecommons.org/licenses/by-nc-nd/3.0/legalcode}\\
	For all other uses including commercial, contact the owner/author(s).
\end{figure}


\section{Introduction}
\label{sec:introduction}

Finding the three-dimensional orientation of a visible rectangle in perspective, such as in optical glyph pose determination, is a challenging task.\cite{Bujnak2008,Garrido-Jurado2014ArUco} Conventional methods typically rely on iterative techniques, such as the coplanar POSIt algorithm,\cite{Oberkampf1996CoplanarPOSIT} which can produce results with significant noise.

These iterative methods often require initial estimations and are sensitive to the quality of input data. Consequently, their performance may be compromised in cases of poor image quality, insufficient feature detection, or extreme perspective distortions. Additionally, such techniques may not be suitable for real-time applications, as they can be computationally expensive and may exhibit latency issues.

Alternative approaches, like homography-based methods, can also be employed for pose estimation. However, they may suffer from limitations when dealing with extreme viewing angles of fish-eye lenses, which introduce significant image distortions. Furthermore, homography-based methods usually require prior rectification of input images, which adds additional layer of complexity.

In light of these challenges, there is a need for a more robust and efficient approach to determine the three-dimensional orientation of visible rectangle glyph in perspective. The ideal method should be capable of handling extreme field of view angles and distortions, while maintaining low latency and high accuracy, particularly for real-time applications and embedded systems with limited computational power.

\paragraph{Note}
If one is not familiar with the topic or terms, appending \vnameref{sec:appendix} gives brief introduction.
\\

This paper addresses the aforementioned issues by introducing a visual-sphere perspective model.\cite{Fober2020VisualSphere} The proposed algorithm calculates the pose matrix and camera position using a closed-form geometric algebra solution to determine spherical vanishing points, which directly map to the axes of the pose matrix. By its nature, the process is limited to coplanar parallel lines. From the pose matrix, the position of the camera and the visible points are reconstructed in three-dimensional space. In this approach, the main function used to calculate the spherical vanishing points is a cross vector product between the incidence vectors of glyph corner points.

The process of pose determination through cross product of incidence was first introduced by \noun{F.A. van den Heuvel} in his paper.\cite{Heuvel1997CoplanarLines} The cited process primarily focused on architectural photogrammetry and planar projections. The novel method presented in this paper extends the previous solution to highly distorted views, beyond the 180\textdegree\ limit of rectilinear projection.\cite{Fleck1995PerspectiveWrong} It also introduces a rectification method for unknown lens parameters (e.g., focal length, angle of view) and a simple focal length estimation method for linear projection. Additionally, this paper provides a geometrical explanation of the method using a spherical perspective imaging model as proof of the proposed solutions.

\subsection[Document Structure]{Structure of the Document}

The paper is organized as follows:

\begin{itemize}
	\item[$\blacksquare$] In \textbf{Section 1} \vpageref{sec:introduction}, an introduction to the paper is given. It includes an overview of the document structure and a brief discussion on the naming convention used throughout the paper.
	\item[$\blacksquare$] \textbf{Section 2} \vpageref{sec:pose_determination} dives into the topic of pose determination. It elaborates on the process of obtaining the incidence vector and the pose matrix.
	\item[$\blacksquare$] \textbf{Section 3} \vpageref{sec:reconstruction} deals with the reconstruction of position. Here, the 3D rectification process is explained, followed by the methods to reconstruct the points' positions and the camera position.
	\item[$\blacksquare$] \textbf{Section 4} \vpageref{sec:unknown_rectification} presents a detailed discussion on 2D rectification, explaining how screen positions are converted to 3D vectors and how the rectification matrix is obtained.
	\item[$\blacksquare$] In \textbf{Section 5} \vpageref{sec:focal_calculation}, the process of calculating the focal length is described, including the formula for line intersection.
	\item[$\blacksquare$] \textbf{Section 6} \vpageref{sec:conclusion} concludes the paper, summarizing the key points and findings discussed in the previous sections.
	\item[$\blacksquare$] Following the conclusion, the \textbf{References} section \vpageref{sec:reference} lists all the sources and materials referred to in the paper.
	\item[$\blacksquare$] Finally, \textbf{Appendix} \vpageref{sec:appendix} provides additional information on pose determination and discusses the PnP problem in depth.
	\item[$\blacksquare$] \textbf{Code Listings} at the end of the paper \vpageref{sec:code} include all the relevant codes used or discussed in the paper.
\end{itemize}


\subsection[Naming Convention]{Document Naming Convention}

This document uses the following naming convention:
\begin{itemize}
	\item A left-handed coordinate system is used.
	\item Vectors are presented in column format.
	\item Matrices use row-major order and are denoted as  ``$M_{\text{row}\,\text{col}}$''.
	\item Matrix multiplication is denoted as ``$[\text{column}]_a\cdot[\text{row}]_b=M_{a\,b}$''.
	\item A single bar enclosure ``$|u|$'' represents the absolute value of a scalar.
	\item A single bar enclosure ``$|\vec v|$'' represents the length of a vector.
	\item Vectors with an arithmetic sign, or without, are calculated component-wise to form another vector.
	\item Centered dot ``$\cdot$'' represents the dot product of two vectors.
	\item Square brackets with a comma ``$[f,c]$'' denotes an interval.
	\item Square brackets with blanks ``$[x\ y]$'' denotes a vector or a matrix.
	\item The power of ``$^{-1}$'' implies the reciprocal of the value.
	\item The \emph{QED} symbol ``$\square$'' marks the final result or output.
\end{itemize}
This naming convention simplifies the process of translating formulas into shader code.

\section[Pose Determination]{Pose Determination Using Rectangular Glyphs}
\label{sec:pose_determination}

\noindent To compute the pose matrix $P$ of a projected $\overline{ABCD}$ rectangle, a method employing multiple cross products between rectangle corners can be utilized.\cite{Heuvel1997CoplanarLines} Each visible point has a $\real^3$ incidence vector, derived from a perspective vector map $G$, or camera focal length and sensor size, or angle of view (AOV) $\Omega$. In this solution, the perspective vector map is favored, as it can describe projections extending beyond 180\textdegree\ of view.\cite{Fober2020VisualSphere}

\subsection[Obtaining Incidence Vector]{Converting Picture Coordinates to Incidence Vector}
\label{sub:incidence_vector}

\begin{figure}[h]
	\includegraphics{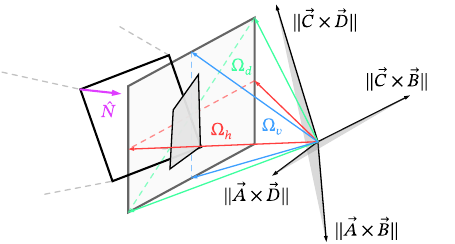}
	\caption[Rectilinear model]
	{Pinhole projection of $\overline{ABCD}$ square (counting clockwise from top-left corner), presenting vectors tangent to projected lines incidence.}
\end{figure}
\noindent
With a perspective vector map texture, the incidence vector $\hat G$ can be easily obtained from $\real^3_{>0}$ pixel values mapped to the $[-1,1]^3$ range, read from the texture at the given picture plane coordinates $\vec f\in[0,1]^2$.

In the case of rectilinear projection, knowing the AOV (denoted as $\Omega$) is sufficient to calculate the incidence vector $\vec G$ from the picture plane coordinates $\vec f$. The following equations describe the process. Additionally, if only the focal length and sensor size are known, $\Omega$ can be computed using the inverse tangent function for rectilinear projection case.

\begin{equation}
	\begin{bmatrix}
		\vec G_x \\
		\vec G_y \\
		\vec G_z
	\end{bmatrix}
	=
	\begin{cases}
		\begin{bmatrix}
			2\vec f_s-1 \\
			(2\vec f_t-1)\div a \\
			\cot\frac{\Omega_h}{2}
		\end{bmatrix}, & \text{if horizontal }\Omega
	\smallskip \\
		\begin{bmatrix}
			a(2\vec f_s-1)\div\sqrt{a^2+1} \\
			(2\vec f_t-1)\div\sqrt{a^2+1} \\
			\cot\frac{\Omega_d}{2}
		\end{bmatrix}, & \text{if diagonal }\Omega
	\smallskip \\
		\begin{bmatrix}
			a(2\vec f_s-1) \\
			2\vec f_t-1 \\
			\cot\frac{\Omega_v}{2}
		\end{bmatrix}, & \text{if vertical }\Omega
	\end{cases}
\end{equation}
a simple algorithm for mapping of texture coordinates $\vec f\in[0,1]^2$ to incident vector $\vec G\in\real^3$ in rectilinear projection. Here $a$ represents picture aspect-ratio and $\Omega$ is the angle of view (aka FOV). Here maximum $\Omega_d<180\degree$.
Same algorithm can be expressed as GLSL function, seen in listing \vref{lst:incident}.

\subsection[Obtaining Pose Matrix]{Estimating Pose Matrix from Incident Vectors}
\label{sub:pose_matrix}

Given the incidence vectors in camera-space for each corner of the projected rectangle, the pose matrix $P$ can be calculated as follows.
\begin{gather}
	\begin{subequations}
		\begin{align}
			\hat X &=
				\Vert
					(\hat A\cross\hat B)
					\cross
					(\hat C\cross\hat D)
				\Vert \\
			\hat Y &=
				\Vert
					(\hat A\cross\hat D)
					\cross
					(\hat C\cross\hat B)
				\Vert
		\end{align}
	\end{subequations}
	\\
	\begin{align}
		\hat Z &= \hat X \cross \hat Y
		\equiv \hat N \\
		P &=
		\begin{bmatrix}
			\hat X_1 \; \hat X_2 \; \hat X_3 \\
			\hat Y_1 \; \hat Y_2 \; \hat Y_3 \\
			\hat Z_1 \; \hat Z_2 \; \hat Z_3 \\
		\end{bmatrix} \qed
	\end{align}
\end{gather}
here, the vectors $\hat X$ and $\hat Y$ point towards two spherical vanishing points. The plane formed by these two vectors is parallel to the plane of the projected rectangle, and therefore, their cross product yields the rectangle's normal vector, $\hat N$. The same algorithm can be expressed as a GLSL function, as demonstrated in listing \vref{lst:pose_matrix}.

\begin{figure}[h]
	\subfloat[short X]
	[Here, vector $\hat X$ points to the spherical vanishing point -- the intersection of great circles $\overline{AB}$ and $\overline{CD}$. Vector $\hat X$ is derived from the normalized cross product between vectors tangent to these two great circles.]
	{\includegraphics{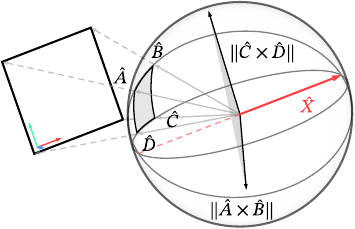}}
\bigskip \\
	\subfloat[short Y]
	[Here, vector $\hat Y$ points to another spherical vanishing point -- the intersection of great circles $\overline{AD}$ and $\overline{CB}$. Vector $\hat Y$ is derived from the normalized cross product between vectors tangent to these two great circles.]
	{\includegraphics{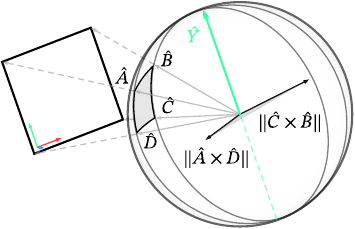}}

	\caption[Pose matrix from vanishing points]
	{Illustrating spherical vanishing points as pose matrix component vectors $P_{\hat 1}$ and $P_{\hat 2}$.}
\end{figure}

\begin{thm}
	The angle between $\hat X$ and $\hat Y$ is equivalent to the angle between corresponding sides of the projected figure (rectangle/square or parallelogram).
\end{thm}
\begin{example*}
	Let us define $\theta$ as the angle between vectors $\hat X$ and $\hat Y$, and $\alpha$ as the angle between the visible corresponding sides of the figure. For a parallelogram, the angle $\theta = \alpha$, and for a rectangle, $\theta$ and $\alpha$ both equal $90\degree$.
\end{example*}
\begin{proof}
	Vectors $\hat X$ and $\hat Y$ point to the vanishing points of parallelogram or rectangle sides. Therefore, the plane on which both vectors lie is parallel to the plane of the projected figure. Pointing to the same vanishing points makes vectors $\hat X$ and $\hat Y$ similar to the corresponding sides of the projected figure, thus having the same angle in between.
\end{proof}

\section[Reconstruction of Position]{Visual Space Position Reconstruction}
\label{sec:reconstruction}

Given the pose matrix and fiducial target dimensions, the position of the camera relative to the target points can be calculated (and vice versa).

\subsection[3D Rectification]{3D Rectification by Plane Intersection}
\label{sub:plane_intersection}

With the pose matrix of the projected figure, the normal vector of the figure's plane can be extracted from the third component of the pose matrix. Points can be then extended to the intersection point with the figure's plane using ratio of the dot products.

\begin{equation}
	\vec A' = \frac{\hat C\cdot\hat N}{\hat A\cdot\hat N}\hat A \qquad
	\vec B' = \frac{\hat C\cdot\hat N}{\hat B\cdot\hat N}\hat B \qquad
	\vec D' = \frac{\hat C\cdot\hat N}{\hat D\cdot\hat N}\hat D
\end{equation}
Equations for extending points $\hat A$, $\hat B$, and $\hat D$ to a plane, with $\hat C$ in the numerator as a reference point lying on the plane's surface.

\begin{align}
	\begin{split}
		\frac{\vec C\cdot\vec N} {\vec A\cdot\vec N}\vec A
		&= \frac{|\vec C|\cancel{|\vec N|}\cos\gamma} {|\vec A| \cancel{|\vec N|}\cos\alpha}\vec A
		\\
		&= \pm\frac
		{\cancel{|\vec C|}
			\frac{h_1}
			{\cancel{|\vec C|}}
		}
		{\cancel{|\vec A|}
			\frac{h_2}
			{\cancel{|\vec A|}}
		} \vec A
		= \pm\frac{h_1}{h_2} \vec A
		= \pm\frac{|\vec A'|}{|\vec A|} \vec A
		= \vec A'
	\end{split}
\end{align}
The sign and length of the normal vector $\hat N$ cancel out, as do the lengths of vectors $\vec A$ and $\vec C$, yielding the proportion of distances $h$ to the plane. This three-dimensional rectification by plane intersection can be expressed as a GLSL function, as shown in listing \vref{lst:plane_intersection}.

\begin{figure}[H]
	\includegraphics{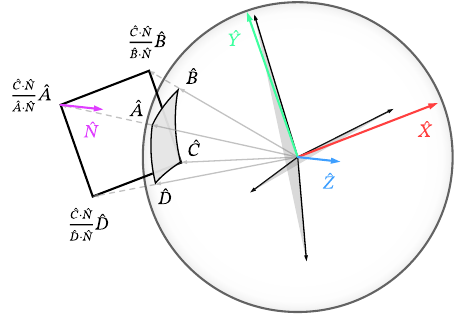}
	\caption[Three-dimensional rectification]
	{
		Illustrating 3D rectification of $\overline{ABCD}$ quad, where normal $\hat N\equiv\hat Z$. Here, pose matrix vector components are represented by
		$\hat X =
		\Vert(\hat A\cross\hat B)
		\cross
		(\hat C\cross\hat D)\Vert$,
		$\hat Y =
		\Vert(\hat A\cross\hat D)
		\cross
		(\hat C\cross\hat B)\Vert$ and
		$\hat Z=\hat X \cross \hat Y$.
	}
\end{figure}

\subsection[Reconstruction of Points Position]{Full 3D Reconstruction of Target Position}
\label{sub:reconstruction}

Given the rectified points $\vec A'$, $\vec B'$, $\hat C$, and $\vec D'$ of the fiducial target with known dimensions, the camera-space 3D position can be reconstructed through simple scalar multiplication.

\begin{align}
	u &= \frac{a}{|\vec B'-\vec A'|} = \frac{b}{|\hat C-\vec B'|} = \frac{c}{|\vec D'-\hat C|} = \frac{d}{|\vec A'-\vec D'|}
	\\
	\vec A'' &= u\vec A' \qquad
	\vec B''  = u\vec B' \qquad
	\vec C'   = u\hat C  \qquad
	\vec D''  = u\vec D'
\end{align}
Where $u$ is the vector scalar of rectified projection points to known size, and $a,b,c,d$ are fiducial target sides length. Vectors $\vec A'', \vec B'', \vec C', \vec D''$ represent reconstructed position of fiducial target in camera-space. The same process can be expressed as a GLSL function, as shown in listing \vref{lst:marker_position}.

\subsection[Reconstruction of Camera Position]{Full 3D Reconstruction of Camera Position}
\label{sub:camera_reconstruction}

The camera's orientation in relation to the fiducial target can be obtained from the pose matrix $P$ and the reconstructed points' positions.

\begin{equation}
	\vec O =
		-\begin{bmatrix}
			\vec D''_x \\
			\vec D''_y \\
			\vec D''_z
		\end{bmatrix}
		\begin{bmatrix}
			P_{11} & P_{12} & P_{13} \\
			P_{21} & P_{22} & P_{23} \\
			P_{31} & P_{32} & P_{33}
		\end{bmatrix}
\end{equation}
Where $\vec D''$ is the fiducial target's origin point (left bottom corner of the target) and $\vec O$ represents the camera's position in the target's space. The same process can be expressed as a GLSL function, as shown in listing \vref{lst:camera_position}.

\section[2D Rectification]{2D Rectification with Unknown Lens Parameters}
\label{sec:unknown_rectification}

Using the visual-sphere vanishing points method, it is possible to rectify a visible quad, seen in rectilinear perspective, without knowing camera-lens parameters, such as focal length or angle of view. The result is a perspective-correct 2D position of a rectangle, albeit with an inaccurate aspect ratio. If the aspect ratio is a fiducial feature, the rectification matrix can be adjusted to compensate for this correction.

\subsection[Screen Position to 3D Vector]{Converting 2D Screen Position to Three Dimensions}
\label{sub:screen_to_3d}

To incorporate the visual-sphere rectification method, the corner's incident 3D vector must be obtained from the 2D screen position. In the best-case scenario, the $z$ value is simply the focal length (for rectilinear projections). However, most of the time, the focal length is either unknown, inaccurate, or not relative to the picture size. In this method, the $z$ distance is substituted with a value of 1 (later in section \vref{sec:focal_calculation}, a calculation method for the focal length is presented).
\begin{equation}
	\vec A' = \begin{bmatrix} \vec A_x \\ \vec A_y \\ 1 \end{bmatrix} \quad
	\vec B' = \begin{bmatrix} \vec B_x \\ \vec B_y \\ 1 \end{bmatrix} \quad
	\vec C' = \begin{bmatrix} \vec C_x \\ \vec C_y \\ 1 \end{bmatrix} \quad
	\vec D' = \begin{bmatrix} \vec D_x \\ \vec D_y \\ 1 \end{bmatrix}
\end{equation}
Where $\vec A,\vec B,\vec C,\vec D \in \real^2 \mapsto \vec A',\vec B',\vec C',\vec D' \in \real^3$ with the origin located at the lens's optical axis.

\subsection[Rectification Matrix]{Rectification Matrix Calculation}
\label{sub:rectification_matrix}

The rectification matrix $R$ is constructed in the same way as the pose matrix $P$ (see subsection \vref{sub:pose_matrix}), but it does not represent orientation.
\begin{gather}
	\begin{subequations}
		\begin{align}
			\hat X &= \Vert(\vec A' \cross \vec B') \cross (\vec C' \cross \vec D')\Vert \\
			\hat Y &= \Vert(\vec A' \cross \vec D') \cross (\vec C' \cross \vec B')\Vert
		\end{align}
	\end{subequations}
	\\
	\begin{align}
		\hat Z &= \Vert\hat Y \cross \hat X\Vert \\
		R &= \begin{bmatrix}
			\hat X_x & \hat X_y & \hat X_z \\
			\hat Y_x & \hat Y_y & \hat Y_z \\
			\hat Z_x & \hat Z_y & \hat Z_z
		\end{bmatrix}
	\end{align}
\end{gather}

\subsubsection[Rectification Process]{Rectification Rotation Process}

The rectification process involves rotation by the rectification matrix $R$ and division of the rotated vector $x,y$ components by the rotated $z$ component. Division places all vectors on a common plane at distance $z=1$.
\begin{equation}
	\vec A'' =
	\begin{bmatrix}
		\frac{\vec A'\cdot \hat X}{\vec A'\cdot \hat Z} \vspace{0.4em}\\
		\frac{\vec A'\cdot \hat Y}{\vec A'\cdot \hat Z}
	\end{bmatrix} \quad
	\vec B'' =
	\begin{bmatrix}
		\frac{\vec B'\cdot \hat X}{\vec B'\cdot \hat Z} \vspace{0.4em}\\
		\frac{\vec B'\cdot \hat Y}{\vec B'\cdot \hat Z}
	\end{bmatrix} \quad
	\vec C'' =
	\begin{bmatrix}
		\frac{\vec C'\cdot \hat X}{\vec C'\cdot \hat Z} \vspace{0.4em}\\
		\frac{\vec C'\cdot \hat Y}{\vec C'\cdot \hat Z}
	\end{bmatrix} \quad
	\vec D'' =
	\begin{bmatrix}
		\frac{\vec D'\cdot \hat X}{\vec D'\cdot \hat Z} \vspace{0.4em}\\
		\frac{\vec D'\cdot \hat Y}{\vec D'\cdot \hat Z}
	\end{bmatrix}
\end{equation}
Where $\vec A'', \vec B'', \vec C'', \vec D'' \in \real^2$ are the rectified versions of vectors $\vec A, \vec B, \vec C, \vec D \in \real^2$.

\paragraph{Aspect Ratio Correction}
can be calculated once for a given lens and combined with each $\hat X$ matrix vector component.
\begin{equation}
	\vec X' = \frac{r|\vec D''-\vec A''|}{|\vec B''-\vec A''|}
	\begin{bmatrix}
		\hat X_x \\ \hat X_y \\ \hat X_z
	\end{bmatrix}
\end{equation}
Where $r$ is the known aspect ratio of the visible rectangle (e.g., for a $4\times3$ aspect, $r=\nicefrac{4}{3}$).

\paragraph{Normalization}
of the rectified coordinates $\vec I$ to $[0,1]^2$ range can be achieved with subtraction and division by opposite, rectified corner points $\vec D''$ and $\vec B''$.
\begin{equation}
	\vec I' = \frac{\vec I-\vec D''}{\vec B''-\vec D''}
\end{equation}

\paragraph{Rectification of Texture Coordinates} requires the transposed matrix $R^T$, as screen texture coordinates are mapped to visible quad coordinates. The non-transposed $R$-matrix rectification maps visible quad corners to screen coordinates.

\section{Calculating Focal Length}
\label{sec:focal_calculation}

Focal length of an unknown rectilinear lens can be calculated from $\hat X$ and $\hat Y$ vectors of the rectification matrix $R$. As they point to two perpendicular vanishing points $\vec V_X, \vec V_Y\in\real^2$ of rectangular target. When pose matrix vectors $\hat X, \hat Y \in P$ are scaled to intersection with the view plane (see formula in subsection \vref{sub:plane_intersection}), their $z$ component is equal to the focal length $f$.
\begin{equation}
	P_{\hat X}\cdot P_{\hat Y}
	= |P_{\hat X}||P_{\hat Y}|\ \cancelto{0}{\cos(\nicefrac{\pi}{2})}\quad
	= 0
\end{equation}
Dot product of pose matrix $P$ vectors $\hat X$, $\hat Y$ equals zero, as they are perpendicular ($P_{\hat X}\ \bot\ P_{\hat Y}$), therefore equation can be rewritten as follows.
\begin{gather}
	\begin{align}
		\begin{split}
			P_{\hat X_x}P_{\hat Y_x}+P_{\hat X_y}P_{\hat Y_y}+P_{\hat X_z}P_{\hat Y_z} &= 0 \\
			P_{\hat X_x}P_{\hat Y_x}+P_{\hat X_y}P_{\hat Y_y} &= -P_{\hat X_z}P_{\hat Y_z}
		\end{split}
		\\
		\vec X'_x\vec Y'_x+\vec X'_y\vec Y'_y &= \pm f^2
	\end{align}
\end{gather}
Here $f$ is the focal length, $\hat X$ and $\hat Y$ are the component vectors of the pose matrix $P$ (see subsection \vref{sub:pose_matrix}), while $\hat X'$ and $\hat Y'$ are $\real^3$ vectors scaled to intersection with the view plane at the $\real^2$ vanishing points $\vec V_{X}$ and $\vec V_{Y}$.

\begin{figure}[h]
	\includegraphics{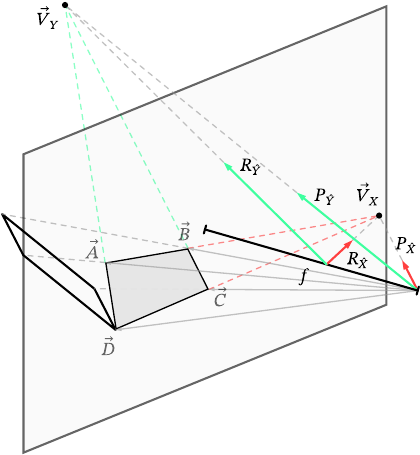}
	\caption[Focal length and matrix]
	{
		Presenting the correlation between focal length $f$, vanishing points $\vec V_X$, $\vec V_Y$ and pose matrix components $\hat X, \hat Y \in P$ with rectification matrix components $\hat X, \hat Y \in R$.
		$P_{\hat X}\ \bot\ P_{\hat Y}$, while $R_{\hat X}\ \cancel{\bot}\ R_{\hat Y}$.
	}
\end{figure}

\begin{equation}
	\vec V = \frac{f}{\hat X_z} \begin{bmatrix}\hat X_x \\ \hat X_y\end{bmatrix} \qed
\end{equation}
Where $\vec V\in\real^2$ is the vanishing point and $\hat X\in\real^3$ represents one of the rectification matrix vectors. If $\hat X_z=0$, the lines are parallel and do not form a vanishing point.
Given that the initial focal distance $f$ in rectification matrix $R$ is equal to one, the formula can be rewritten as follows.
\begin{align}
	\vec V_X &= \begin{bmatrix} R_{\hat X_x} \\ R_{\hat X_y} \end{bmatrix}\div R_{\hat X_z}
	\qquad
	\vec V_Y = \begin{bmatrix} R_{\hat Y_x} \\ R_{\hat Y_y} \end{bmatrix}\div R_{\hat Y_z} \qed
	\\
	\begin{split}
		f &= \sqrt{|\vec V_X\cdot\vec V_Y|} \\
		  &= \sqrt{\left|\frac
		  	{ R_{\hat X_x} R_{\hat Y_x} + R_{\hat X_y} R_{\hat Y_y} }
		  	{ R_{\hat X_z} R_{\hat Y_z} }
		  \right|} \qed
	\end{split}
\end{align}
Where $\vec V_X, \vec V_Y \in \real^2$ are the horizontal and vertical vanishing points, respectively. $f$ is the focal length calculated from rectification matrix $R$ vectors $\hat X$ and $\hat Y$. For the rectification matrix formula, see subsection \vref{sub:rectification_matrix}.

\subsection{Line Intersection Formula}
\label{sub:line_intersection}

The equation for the vanishing point $\vec V$ can be utilized as a generic 2D line-line intersection formula. When the denominator is equal to zero ($\hat X_z$ or $\hat Y_z$), the lines are parallel.

\begin{figure}[h]
	\includegraphics{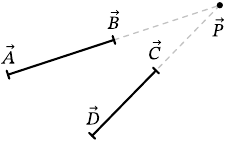}
	\caption[Line-line intersection]
	{2D line-line intersection model, where vector $\vec P$ represents the intersection point of lines $\overline{AB}$ and $\overline{CD}$.}
\end{figure}
\begin{align}
	\vec A' = \begin{bmatrix}\vec A_x \\ \vec A_y \\ 1\end{bmatrix} \quad
	\vec B' &= \begin{bmatrix}\vec B_x \\ \vec B_y \\ 1\end{bmatrix} \quad
	\vec C' = \begin{bmatrix}\vec C_x \\ \vec C_y \\ 1\end{bmatrix} \quad
	\vec D' = \begin{bmatrix}\vec D_x \\ \vec D_y \\ 1\end{bmatrix} \\
	\vec X &= (\vec A'\cross\vec B')\cross(\vec C'\cross\vec D') \\
	\vec P &= \begin{bmatrix}\vec X_x \\ \vec X_y\end{bmatrix}\div \vec X_z \qed
\end{align}
Where $\vec P\in\real^2$ is the intersection point of lines $\overline{AB}$ and $\overline{CD}$.

\section{Conclusion}
\label{sec:conclusion}

In this paper, I have demonstrated that the orientation matrix (a pose matrix) can be directly evaluated in a single iteration from the spherical vanishing points of coplanar parallel lines forming a rectangle. Furthermore, the three-dimensional position of this visible rectangle can be reconstructed in a direct and finite manner. Such a process could be easily integrated into a hardware solution for pose calculation and position reconstruction of square markers. The close relation with the visual-sphere perspective extends the use of this algorithm to wide-angle lenses (e.g., fish-eye lenses), the properties of which exceed the geometrical limits for field-of-view of standard linear perspective projection.\cite{Fleck1995PerspectiveWrong}

Furthermore, using the same formula, I have presented a generic perspective rectification method. This method solves the rectification problem for unknown rectilinear lens parameters. I have also shown that the formula for focal length estimation can function, in another form, as a generic solution for 2D line-line intersection point calculation.

Despite these promising results, the method proposed in this paper assumes ideal conditions and further research is needed to validate its performance in real-world scenarios, where factors such as parallax lens distortion, noise, and lighting conditions may affect the results.

This set of solutions is well-suited for embedded computer vision tasks that require high performance and precision, making it ideal for hardware integration. Future work could explore the potential of integrating these methods into various computer vision systems and applications, and further optimize the algorithms for real-time processing.



\vfill
\label{sec:reference}
\bibliographystyle{ACM-Reference-Format}
\bibliography{bibliography} 


\begin{thebibliography}{6}


\ifx \showCODEN    \undefined \def \showCODEN     #1{\unskip}     \fi
\ifx \showDOI      \undefined \def \showDOI       #1{#1}\fi
\ifx \showISBNx    \undefined \def \showISBNx     #1{\unskip}     \fi
\ifx \showISBNxiii \undefined \def \showISBNxiii  #1{\unskip}     \fi
\ifx \showISSN     \undefined \def \showISSN      #1{\unskip}     \fi
\ifx \showLCCN     \undefined \def \showLCCN      #1{\unskip}     \fi
\ifx \shownote     \undefined \def \shownote      #1{#1}          \fi
\ifx \showarticletitle \undefined \def \showarticletitle #1{#1}   \fi
\ifx \showURL      \undefined \def \showURL       {\relax}        \fi
\providecommand\bibfield[2]{#2}
\providecommand\bibinfo[2]{#2}
\providecommand\natexlab[1]{#1}
\providecommand\showeprint[2][]{arXiv:#2}

\bibitem[Bujnak et~al\mbox{.}(2008)]%
        {Bujnak2008}
\bibfield{author}{\bibinfo{person}{Martin Bujnak}, \bibinfo{person}{Zuzana
  Kukelova}, {and} \bibinfo{person}{Tomas Pajdla}.}
  \bibinfo{year}{2008}\natexlab{}.
\newblock \showarticletitle{A general solution to the P4P problem for camera
  with unknown focal length}. In \bibinfo{booktitle}{\emph{2008 {IEEE}
  Conference on Computer Vision and Pattern Recognition}}.
  \bibinfo{publisher}{{IEEE}}.
\newblock
\urldef\tempurl%
\url{https://doi.org/10.1109/cvpr.2008.4587793}
\showDOI{\tempurl}


\bibitem[Fleck(1995)]%
        {Fleck1995PerspectiveWrong}
\bibfield{author}{\bibinfo{person}{Margaret~M. Fleck}.}
  \bibinfo{year}{1995}\natexlab{}.
\newblock \showarticletitle{Perspective projection: the wrong imaging model}.
\newblock \bibinfo{journal}{\emph{Department of Computer Science, University of
  Iowa}} (\bibinfo{date}{Jan.} \bibinfo{year}{1995}), \bibinfo{pages}{1--27}.
\newblock
\urldef\tempurl%
\url{https://citeseerx.ist.psu.edu/document?repid=rep1&type=pdf&doi=7a80a9c9068a6a5ea5836a56fd440f2477cda17c}
\showURL{%
\tempurl}


\bibitem[Fober(2020)]%
        {Fober2020VisualSphere}
\bibfield{author}{\bibinfo{person}{Jakub~Maksymilian Fober}.}
  \bibinfo{year}{2020}\natexlab{}.
\newblock \showarticletitle{Perspective picture from Visual Sphere: A new
  approach to image rasterization}.
\newblock \bibinfo{journal}{\emph{ArXiv}} (\bibinfo{year}{2020}).
\newblock
\urldef\tempurl%
\url{https://doi.org/10.48550/arXiv.2003.10558}
\showDOI{\tempurl}


\bibitem[Garrido-Jurado et~al\mbox{.}(2014)]%
        {Garrido-Jurado2014ArUco}
\bibfield{author}{\bibinfo{person}{S. Garrido-Jurado}, \bibinfo{person}{R.
  Mu{\~n}oz-Salinas}, \bibinfo{person}{F.~J. Madrid-Cuevas}, {and}
  \bibinfo{person}{M.~J. Mar{\'i}n-Jim{\'e}nez}.}
  \bibinfo{year}{2014}\natexlab{}.
\newblock \showarticletitle{Automatic generation and detection of highly
  reliable fiducial markers under occlusion}.
\newblock \bibinfo{journal}{\emph{Pattern Recognition}} \bibinfo{volume}{47},
  \bibinfo{number}{6} (\bibinfo{date}{June} \bibinfo{year}{2014}),
  \bibinfo{pages}{2280--2292}.
\newblock
\urldef\tempurl%
\url{https://doi.org/10.1016/j.patcog.2014.01.005}
\showDOI{\tempurl}


\bibitem[Heuvel(1997)]%
        {Heuvel1997CoplanarLines}
\bibfield{author}{\bibinfo{person}{Frank A. Van~Den Heuvel}.}
  \bibinfo{year}{1997}\natexlab{}.
\newblock \showarticletitle{Exterior orientation using coplanar parallel
  lines}. In \bibinfo{booktitle}{\emph{Proceedings of the Scandinavian
  Conference on Image Analysis}}, Vol.~\bibinfo{volume}{1}.
\newblock
\urldef\tempurl%
\url{https://citeseerx.ist.psu.edu/viewdoc/download?doi=10.1.1.2.1117&rep=rep1&type=pdf}
\showURL{%
\tempurl}


\bibitem[Oberkampf et~al\mbox{.}(1996)]%
        {Oberkampf1996CoplanarPOSIT}
\bibfield{author}{\bibinfo{person}{Denis Oberkampf}, \bibinfo{person}{Daniel~F.
  DeMenthon}, {and} \bibinfo{person}{Larry~S. Davis}.}
  \bibinfo{year}{1996}\natexlab{}.
\newblock \showarticletitle{Iterative Pose Estimation Using Coplanar Feature
  Points}.
\newblock \bibinfo{journal}{\emph{Computer Vision and Image Understanding}}
  \bibinfo{volume}{63}, \bibinfo{number}{3} (\bibinfo{date}{May}
  \bibinfo{year}{1996}), \bibinfo{pages}{495--511}.
\newblock
\urldef\tempurl%
\url{https://doi.org/10.1006/cviu.1996.0037}
\showDOI{\tempurl}


\end{thebibliography}


\pagebreak
\appendix
\section{Pose Determination}
\label{sec:appendix}

Pose determination is a common technique in computer vision used to reproduce physical space from a two-dimensional symbolic picture. It often involves registering the position of fiducial markers, such as color points in movie special effects or binary square fiducial markers (optical glyphs) like ArUco markers for other purposes \cite{Garrido-Jurado2014ArUco}. While movie special effects focus on the P$n$P problem, a more constrained fiducial environment offers simpler and more repeatable methods of 3D reconstruction. In the case of square fiducial markers, the number of corner points, perpendicularity of edges, and opposite parallelism can be treated as fiducial features.

To benefit from such constants, basic principles must be altered, like the perspective projection model. Two vanishing points of a rectangle visible in perspective will point to two component vectors of the pose matrix. However, in the case of rectilinear projection, the vanishing point position can easily approach infinity when one of its edges is nearly parallel to the projection plane. Such large numbers are undesirable in computational geometry, as they are prone to reach precision limits.

An alternative approach to vanishing points involves visual-sphere perspective \cite{Fober2020VisualSphere}, where vanishing points are formed by the intersection of great circles.

\begin{rem}
	Spherical perspective geometry can be defined through normalized Euclidean $\real^3$ vectors, which avoid the spherical coordinate $\real^2$ system and simplify calculations.
\end{rem}

\subsection[On PnP Problem]
{On Coplanar-\textit{n}-Point Problem}
\label{sub:pnp_problem}

\begin{problem*}
	What is the smallest number of points required for complete pose determination?
	Let us consider only a case where back-facing does not occur and the image cannot be mirrored.
	\begin{example*}
		Photographing binary fiducial markers on solid planar surfaces.
	\end{example*}
\end{problem*}
\begin{thm}
	The minimum number of points for pose determination is four with an additional fiducial cue. If only simple points are considered, the fiducial cue becomes the fifth point.
\end{thm}

\begin{figure}[h]
	\includegraphics{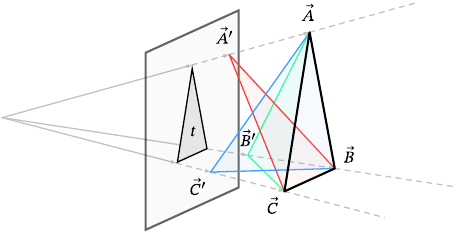}
	\caption[Poses in P3P problem]
	{Visualization of the P3P problem for projected equilateral triangle $t$, where $\overline{ABC}\sim\overline{A'BC}\sim\overline{AB'C}\sim\overline{ABC'}$, yielding four possible normal directions.}
	\label{fig:p3p problem}
\end{figure}

\paragraph[P3P]{P3P}\label{par:p3p} The problem for a projected equilateral triangle in perspective yields four possible normal vectors with three possible symmetry rotations, giving a total of twelve possible pose matrices (see Figure \vref{fig:p3p problem}). Symmetry rotation can be resolved with additional cues enabling point sorting. \qed

\paragraph[P4P]{P4P}\label{par:p4p} The problem for a projected square in perspective has a single possible normal direction and four possible symmetry rotations, giving a total of four possible pose matrices. Sorting visible points with additional fiducial cues can limit the number of possible symmetry rotations. Such cues can be derived from fiducial markers, using color, shape, size, or other unique features or known conditions. \qed

\paragraph[P5P]{P5P}\label{par:p5p} The problem for a projected square in perspective with an additional point on one of the square's sides yields a single possible pose matrix. In such a case, the fifth point is used as a fiducial marker for sorting the other four points.
The fifth point in such a configuration can be simply extracted by clustering, alternatively by measuring collinearity, or in the case of visual sphere perspective, coplanarity. \qed
\begin{rem}
	In the visual sphere perspective, $[-1,1]^3$ points $\hat A, \hat B, \hat C$ belong to a single great circle if $\hat A \times \hat B \cdot \hat C = 0$, as the cosine of 90\textdegree\ is equal to zero.\footnote{This algorithm is known as the \emph{triple product}.}
\end{rem}


\onecolumn
\pdfbookmark{Code Listings}{bm:listings}
\label{sec:code}

\begin{lstlisting}[
	firstline=7, % Bypass licence notification (repeated)
	caption={[Rectilinear incident vector]
		Function mapping texture coordinates $\vec f\in[0,1]^2$, to incident vector $\vec G\in\real^3$ in GLSL, for rectilinear view.},
	label={lst:incident}
]
vec3 getIncident(vec2 tex_coord, float aspect, float fov, uint fov_type)
{
	vec3 incident = vec3(
		2.0*tex_coord-1.0, // map to range [-1,1]
		1.0/tan(0.5*radians(fov) )
	);
	switch (fov_type)
	{
		default: // horizontal FOV (type 1)
			incident.y /= aspect;
			break;
		case 2: // diagonal FOV (type 2)
			incident.xy /= length(vec2(aspect, 1.0) );
		case 3: // vertical FOV (type 3) and type 2 final step
			incident.x *= aspect;
			break;
	}

	return incident;
}
\end{lstlisting}

\begin{lstlisting}[
	firstline=7, % Bypass licence notification (repeated)
	caption={[Pose matrix]
		Function for pose matrix $P$ in GLSL, where matrix \textbit{``quad''} represents four $\real^3$ incident-vectors of projected rectangle corners.},
	label={lst:pose_matrix}
]
mat3 getPose(mat4x3 quad)
{
	mat3 Pose;
	// X axis vector
	Pose[0] = cross(
		cross(quad[0], quad[1]),
		cross(quad[2], quad[3])
	);
	// Y axis vector
	Pose[1] = cross(
		cross(quad[0], quad[3]),
		cross(quad[2], quad[1])
	);
	// Z axis vector
	Pose[2] = cross(Pose[0], Pose[1]);
	// Normalize matrix
	Pose[0] = normalize(Pose[0]); // X axis
	Pose[1] = normalize(Pose[1]); // Y axis
	Pose[2] = normalize(Pose[2]); // Z axis

	return Pose;
}
\end{lstlisting}

\begin{lstlisting}[
	firstline=7, % Bypass licence notification (repeated)
	caption={[Plane intersection function]
		Functions for vector--plane intersection in GLSL.},
	label={lst:plane_intersection}
]
vec3 toPlane(vec3 vector, vec3 normal, vec3 plane_pt)
{ return dot(plane_pt, normal)/dot(vector, normal)*vector; }

mat4x3 toPlane(mat4x3 quad, mat3 Pose, vec3 plane_pt)
{
	float numerator = dot(plane_pt, Pose[2]);
	for (int i=0; i<4; i++)
		quad[i] *= numerator/dot(quad[i], Pose[2]);

	return quad;
}
\end{lstlisting}

\begin{lstlisting}[
	firstline=7, % Bypass licence notification (repeated)
	caption={[Marker position reconstruction]
		Fiducial marker points position reconstruction function in GLSL, where matrix \textbit{``figure''} represents rectified figure in camera space.},
	label={lst:marker_position}
]
mat4x3 reconstructByWidth(mat4x3 figure, float width)
{ return width/length(figure[2]-figure[3])*figure; }

mat4x3 reconstructByHeight(mat4x3 figure, float height)
{ return height/length(figure[0]-figure[3])*figure; }
\end{lstlisting}

\begin{lstlisting}[
	firstline=7, % Bypass licence notification (repeated)
	caption={[Camera position reconstruction]
		Camera position and orientation reconstruction function in GLSL.},
	label={lst:camera_position}
]
vec3 getCameraPos(mat3 Pose, vec3 marker_origin_pt)
{ return -marker_origin_pt*Pose; }
\end{lstlisting}

\begin{lstlisting}[
	firstline=7, % Bypass licence notification (repeated)
	caption={[2D rectification]
		2D rectification function with unknown camera parameters in GLSL.},
	label={lst:rectification}
]
// 2D rectification function
vec2 rectify(vec2 vector, mat3 rectifyMat)
{
	return vec2(
		dot(vec3(vector, 1.0), rectifyMat[0]),
		dot(vec3(vector, 1.0), rectifyMat[1])
	)/dot(vec3(vector, 1.0), rectifyMat[2]);
}
// Rectification matrix from 2D quad
mat3 getRectifyMatrix(mat4x2 quad2D)
{
	mat3 rectifyMat; // Rectification matrix

	// X vector
	rectifyMat[0] = cross(
		cross(vec3(quad2D[0], 1.0), vec3(quad2D[1], 1.0) ),
		cross(vec3(quad2D[2], 1.0), vec3(quad2D[3], 1.0) )
	);
	// Y vector
	rectifyMat[1] = cross(
		cross(vec3(quad2D[0], 1.0), vec3(quad2D[3], 1.0) ),
		cross(vec3(quad2D[2], 1.0), vec3(quad2D[1], 1.0) )
	);
	// Z vector
	rectifyMat[2] = cross(rectifyMat[1], rectifyMat[0]);
	// Normalize matrix
	rectifyMat[0] = normalize(rectifyMat[0]);
	rectifyMat[1] = normalize(rectifyMat[1]);
	rectifyMat[2] = normalize(rectifyMat[2]);

	return rectifyMat;
}
// Aspect ratio correction
float getAspectDiff(mat3 rectifyMat, mat4x2 quad2D, float aspect)
{
	// Rectify quad for normalization
	quad2D[3] = rectify(quad2D[3], rectifyMat);

	return aspect*
	length(rectify(quad2D[0], rectifyMat)-quad2D[3])/
	length(rectify(quad2D[2], rectifyMat)-quad2D[3]);
}
\end{lstlisting}

\begin{lstlisting}[
	firstline=7, % Bypass licence notification (repeated)
	caption={[Focal length estimation]
		Focal length estimation function with unknown camera parameters in GLSL. See listing \vref{lst:rectification} for rectification matrix function.},
	label={lst:focal}
]
// Focal length from rectification matrix
float getFocalLength(mat3 rectif)
{
	if (rectif[0].z==0.0 || rectif[1].z==0.0) return 0.0; // Planar view, without perspective
	else return sqrt(abs(
		dot(rectif[0].xy, rectif[1].xy)
		/(rectif[0].z*rectif[1].z)
	));
}
// Focal length from matrix X, Y components
float getFocalLength(vec3 X, vec3 Y)
{
	if (X.z==0.0 || Y.z==0.0) return 0.0;
	else return sqrt(abs( dot(X.xy, Y.xy)/(X.z*Y.z) ));
}
// Focal length from visible quad in perspective
float getFocalLength(mat4x2 quad)
{
	// Get quad U axis vector
	vec3 U = cross(
		cross(vec3(quad[0], 1.0), vec3(quad[1], 1.0) ),
		cross(vec3(quad[2], 1.0), vec3(quad[3], 1.0) )
	);
	// Get quad V axis vector
	vec3 V = cross(
		cross(vec3(quad[0], 1.0), vec3(quad[3], 1.0) ),
		cross(vec3(quad[2], 1.0), vec3(quad[1], 1.0) )
	);

	if (U.z==0.0 || V.z==0.0) return 0.0; // Quad not in perspective
	else return sqrt(abs( dot(U.xy, V.xy)/(U.z*V.z) )); // Focal length
}
\end{lstlisting}

\end{document}